%% file: main.tex
\documentclass[10pt,twocolumn,letterpaper]{article}

\usepackage{cvpr}
\input{preamble}

\definecolor{cvprblue}{rgb}{0.21,0.49,0.74}
\usepackage[pagebackref,breaklinks,colorlinks,allcolors=cvprblue]{hyperref}

\def\paperID{*****}
\def\confName{CVPR}
\def\confYear{2026}

\title{TAP-JEPA: Frozen Future-Latent Probing and Two-Stage Score Fusion for EPIC-KITCHENS-100 Action Anticipation}

\author{
Chaoyang Wang$^{1}$, Lexuan Xu$^{1}$\\
$^{1}$Beihang University\\
{\tt\small \{chaoyangwang, xulexuan\}@buaa.edu.cn}
}

\begin{document}
\maketitle
\input{sec/0_abstract}
\input{sec/1_intro}
\input{sec/2_formatting}
\input{sec/3_finalcopy}
\input{sec/4_conclusion}
{
    \small
    \bibliographystyle{unsrt}
    \bibliography{main}
}
\end{document}

%% file: preamble.tex
\usepackage{multirow}
\usepackage{threeparttable}
\usepackage{array}

%% file: sec/0_abstract.tex
\begin{abstract}
This report presents \textbf{TAP-JEPA}, our runner-up submission to the EPIC-KITCHENS-100 (EK-100) Action Anticipation Challenge at EgoVis 2026. The task is to anticipate the next verb, noun, and verb-noun action from an egocentric clip that ends before the target action begins. Instead of fine-tuning a large video backbone, TAP-JEPA builds a compact anticipation model on frozen V-JEPA 2.1 features: a ViT-G/384 encoder extracts visible pre-action tokens, the pretrained latent predictor estimates near-future tokens from the observed context, and both token groups are fused by attentive probes with task-specific queries for verbs, nouns, and action pairs. For the final submission, we expand supervised training with the official training split and most of the validation split, reserving a small subset for sanity checks and qualitative inspection, and adopt a two-stage score fusion that first averages eight independently initialized probe replicas within each epoch and then merges candidates from epochs 12--20 with field-dependent weights. On the official open-testing leaderboard, our \textit{sunshinesky} entry achieves 27.91\% overall action Mean Top-5 Recall (MT5R), ranking second and only 0.04 percentage points behind the top score.
\end{abstract}

%% file: sec/1_intro.tex
\section{Introduction}
\label{sec:intro}

\begin{figure*}[t]
  \centering
  \includegraphics[width=\linewidth]{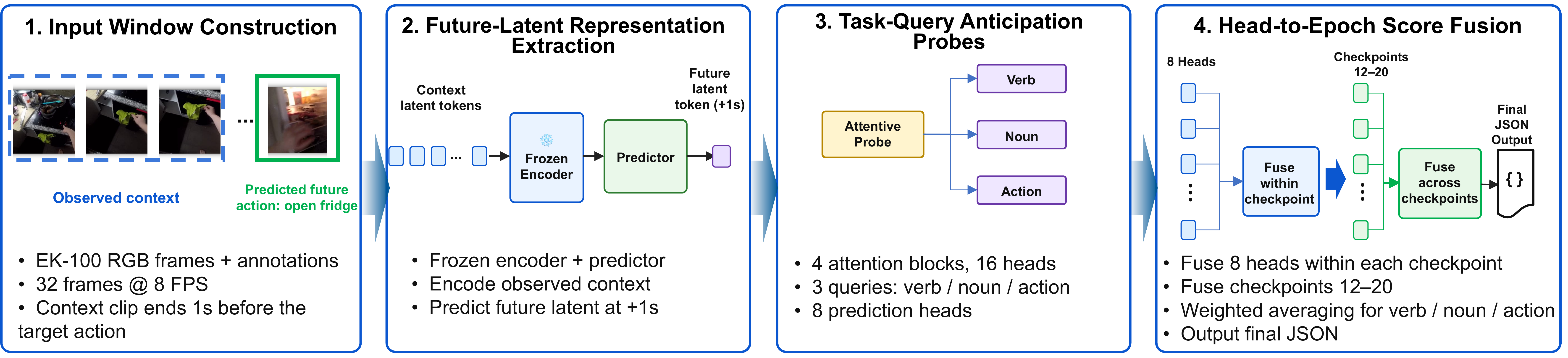}
  \caption{TAP-JEPA inference pipeline. A frozen V-JEPA 2.1 encoder-predictor first turns the observed pre-action clip into visible-context tokens and predicted future-latent tokens. Eight lightweight attentive probe replicas then produce verb, noun, and action scores. The submitted scores are obtained by fusing replicas inside each epoch and subsequently fusing late-epoch candidates from epochs 12-20.}
  \label{fig:method}
\end{figure*}

Action anticipation in egocentric video differs from standard action recognition because the target action has not yet been observed when the prediction is made.
In kitchen recordings, this early-decision requirement is especially challenging: the same hand approach, object layout, or camera motion may lead to several plausible future manipulations.
A model therefore has to exploit weak preparatory evidence, such as hand-object distance, object affordance, scene state, and recent camera motion, rather than relying on the discriminative frames of the action itself.

The EPIC-KITCHENS benchmark provides a natural testbed for this problem.
The original dataset introduced large-scale unscripted egocentric kitchen videos with dense action and object-interaction annotations~\cite{damen2021epickitchens}, and EK-100 further expands the benchmark in scale and annotation coverage~\cite{damen2022rescaling}.
In the EK-100 Action Anticipation Challenge, each input is a pre-action video clip that ends before the start of the target action.
The official anticipation gap is one second, and the system must output the future verb, noun, and verb-noun action.
No frame after the anticipation boundary is used by our inference pipeline.

The evaluation protocol emphasizes class-balanced forecasting rather than only frequent-class accuracy.
EK-100 contains about 100 hours of full-HD egocentric video recorded in 45 kitchens across 4 cities, with 20M frames, 90K action segments, 20K unique narrations, 97 verb classes, and 300 noun classes~\cite{damen2022rescaling}.
Submissions are ranked by Mean Top-5 Recall (MT5R), where Top-5 Recall is first computed per class and then averaged across classes.
The leaderboard also reports performance on unseen participants and tail classes, so a competitive submission must remain stable under both user shift and long-tailed category distributions.

Recent progress in egocentric anticipation further suggests that strong systems should combine robust visual perception, future-aware modeling, and structured reasoning.
Recent challenge-winning reports, such as the Ego4D LTA 2025 champion~\cite{chu2025ego4dlta}, INSIGHT~\cite{chu2026insight}, JFAA~\cite{chu2026jfaa}, and VISTA~\cite{chu2026vista}, show that careful combination of visual features, future-predictive representations, and task-specific probes leads to remarkably strong long- and short-term anticipation performance.

In this context, we propose \textbf{TAP-JEPA}, a frozen future-latent probing approach based on V-JEPA 2.1~\cite{vjepa21}.
Instead of adapting all parameters of a video foundation model, TAP-JEPA keeps the V-JEPA 2.1 ViT-G/384 encoder and predictor fixed and trains only lightweight attentive probes.
The encoder produces latent tokens for the visible context, while the predictor supplies near-future latent tokens without observing future RGB frames.
By passing both token sets to task-query probes, the model combines direct pre-action evidence with a feature-space prediction of the near future.

This design is motivated by two practical considerations.
First, freezing the representation backbone keeps supervised optimization focused on the EK-100 label space and limits overfitting when many classes are rare.
Second, the small trainable probe makes it affordable to introduce ensemble diversity without retraining multiple large backbones.
For the final model, we train eight independently initialized probe replicas and run the optimization for 20 epochs.
Prediction scores are fused in two stages: replica outputs are combined within each epoch, and the resulting epoch-level candidates from epochs 12--20 are averaged with field-specific weights.
This separates intra-checkpoint diversity from late-training checkpoint diversity and allows verb, noun, and action fields to use different score mixtures.

Our final submission, \textit{sunshinesky}, achieves 27.91\% overall action MT5R on the official hidden test set and ranks second in the open-testing phase.
The small gap to the top score shows that frozen future-predictive representations, compact probing, and calibrated two-stage score fusion form an effective recipe for EK-100 action anticipation.

%% file: sec/2_formatting.tex
\section{Method}
\label{sec:method}
Figure~\ref{fig:method} summarizes the TAP-JEPA inference and submission pipeline. The pipeline consists of four stages: input-window construction, frozen future-latent representation extraction, task-query anticipation probing, and two-stage score fusion. We first construct pre-action clips under the official one-second anticipation constraint. The clips are then encoded by a frozen V-JEPA 2.1 encoder-predictor, which yields visible-context tokens and predicted future-latent tokens. Lightweight attentive probe replicas map the concatenated token sequence to verb, noun, and action scores. Finally, the exported challenge file is produced by fusing both probe replicas and late-epoch prediction candidates.

\subsection{Input Window Construction}

We use the official EK-100 RGB frames and annotations~\cite{damen2022rescaling}.
For every target action, the model input is sampled from the interval before the action onset, and the test-time clip ends one second before the action begins.
The final challenge model is trained with the official training split and most of the official validation split.
A small subset is held out only for sanity checks, qualitative inspection, and monitoring the stability of late checkpoints.

Each clip contains 32 RGB frames sampled at 8 FPS.
The frames are resized and cropped to $384\times384$ before being passed to the V-JEPA 2.1 backbone.
During training, we apply the temporal perturbation used in the V-JEPA 2.1 EK-100 anticipation setup~\cite{vjepa21}: the anticipation time is sampled between 0.25 and 1.75 seconds, and the anticipation-point offset is sampled between 0.0 and 0.25 seconds.
This augmentation exposes the probes to nearby pre-action contexts while respecting the rule that target-action frames are not provided as input.
For spatial augmentation, we use random resized crop, RandAugment~\cite{cubuk2020randaugment}, horizontal flip, and random erasing~\cite{zhong2020random}.
Validation and test exports use deterministic resizing and center cropping.

\subsection{Future-Latent Representation Extraction}

The visual representation module is V-JEPA 2.1 ViT-G/384~\cite{bardes2024vjepa,vjepa21,dosovitskiy2021vit}.
V-JEPA 2.1 learns video representations by predicting target features in latent space rather than reconstructing pixels, encouraging the model to capture semantic and temporal regularities in feature space~\cite{bardes2024vjepa}.
V-JEPA 2.1 extends this framework with dense predictive supervision, deep self-supervision over intermediate layers, image/video tokenization, and larger-scale training~\cite{vjepa21}.
These properties are well aligned with egocentric anticipation, where useful evidence may be distributed across hands, manipulated objects, and global scene layout.

We load the pretrained encoder and predictor and freeze all their parameters during supervised training.
Given a pre-action clip, the encoder outputs spatiotemporal tokens for the observed frames.
The predictor then receives the encoder representation together with learnable mask tokens at the future target position and estimates the corresponding future latent representation.
The model never observes the future RGB frame; only the pretrained predictor's feature-space estimate is used.
Following the V-JEPA 2.1 anticipation protocol~\cite{vjepa21}, we concatenate encoder tokens and predictor tokens along the token dimension before downstream classification.

This frozen-latent design makes V-JEPA 2.1 a prior over likely near-future visual states.
It also reduces the number of supervised parameters, which is useful for a long-tailed challenge setting and enables multiple probe replicas to be trained at low additional cost.

\subsection{Task-Query Anticipation Probes}

The trainable module on top of V-JEPA 2.1 is an attentive probe.
Each probe replica follows the V-JEPA 2.1 action-anticipation probe design~\cite{vjepa21} and contains four attention-based blocks with 16 attention heads~\cite{vaswani2017attention}.
A final cross-attention layer uses three learnable task queries, producing separate representations for verb, noun, and action prediction.
The query outputs are passed to three linear classifiers.

The three branches are kept separate because they rely on different types of evidence.
Verb prediction is strongly related to motion and interaction dynamics, noun prediction depends more on object appearance and scene context, and action-pair prediction must combine the two while respecting valid verb-noun compositions.
We optimize all branches with sigmoid focal loss~\cite{lin2017focalloss}, using $\alpha=0.25$ and $\gamma=2.0$ as in the V-JEPA 2.1 setup~\cite{vjepa21}.
The three losses are summed for supervised training.

For the challenge submission, we train eight independently initialized probe replicas on the same frozen token sequence.
Each replica has its own attention blocks, task queries, and classifiers.
Therefore, the term ``head'' in our fusion procedure refers to an independent probe replica rather than a Transformer attention head.
All replicas are trained for 20 epochs with mixed precision~\cite{micikevicius2018mixed}.
This gives a lightweight source of model diversity without fine-tuning or duplicating the large V-JEPA 2.1 backbone.

\subsection{Head-to-Epoch Score Fusion}

TAP-JEPA uses two score-fusion stages.
The first stage aggregates probe replicas within a single checkpoint.
Let $e$ denote the epoch, $h\in\{1,\ldots,8\}$ denote the probe replica, and $f\in\{\mathrm{verb},\mathrm{noun},\mathrm{action}\}$ denote the output field.
For a test timestamp $x$, replica $h$ produces logits $z_{f,e,h}(y\mid x)$ for candidate label $y$.
Since the classifiers are trained with sigmoid focal loss, logits are converted to sigmoid scores before fusion:
\begin{equation}
\begin{aligned}
  s_{f,e}(y\mid x)
  &=
  \sum_{h=1}^{8}
  \alpha_{f,h}\,
  \sigma\!\left(z_{f,e,h}(y\mid x)\right), \\
  \text{s.t.}\quad
  &\sum_{h=1}^{8}\alpha_{f,h}=1,\quad
  \alpha_{f,h}\ge 0 .
\end{aligned}
\label{eq:head_fusion}
\end{equation}
The output of this stage is one fused prediction candidate for each exported epoch.

For verbs and nouns, the scores are expanded to the official 97 verb classes and 300 noun classes.
For action prediction, scores are mapped to valid verb-noun candidates, and the top-100 ranked pairs are written for every test timestamp.
Thus, the action export follows the challenge format while still using the model's action-pair scores.

The second stage fuses candidates across training epochs.
We use the contiguous late-training window from epochs 12-20.
This window is chosen to avoid the early optimization transient while retaining checkpoint diversity near convergence.
The fusion unit is an epoch-level candidate that has already combined all eight probe replicas.
This procedure is related to checkpoint or snapshot ensembling~\cite{huang2017snapshot}, but our candidates come from consecutive late checkpoints rather than a cyclic learning-rate schedule.

Let $e\in\{12,\ldots,20\}$ index the selected epoch-level candidates.
The final score for field $f$ is
\begin{equation}
\begin{aligned}
  S_f(y\mid x)
  &=
  \sum_{e=12}^{20}
  \beta_{f,e}\,
  s_{f,e}(y\mid x), \\
  \text{s.t.}\quad
  &\sum_{e=12}^{20}\beta_{f,e}=1,\quad
  \beta_{f,e}\ge 0 .
\end{aligned}
\label{eq:epoch_fusion}
\end{equation}
The weights are normalized independently for verbs, nouns, and actions.
This is important because the three fields may peak at different checkpoints: verbs favor motion-sensitive cues, nouns favor object-discriminative cues, and action pairs are additionally affected by valid-pair coverage and long-tail imbalance.
The same field-wise fusion rule is applied to all test examples before writing the official action-anticipation JSON file.

%% file: sec/3_finalcopy.tex
\section{Experiments}
\label{sec:experiments}

\begin{table*}[t]
  \centering
  \caption{Official open-testing leaderboard on EK-100 action anticipation for the 2025 Open Testing Phase Task. All entries report test-set Mean Top-5 Recall (MT5R), obtained by averaging class-wise Top-5 Recall over classes appearing in the test set. Score denotes overall action MT5R (O-A). PT, TL, and TD denote the official Supervision Levels Scale (SLS) dimensions: pre-training, training labels, and amount of training data. O, U, and T denote Overall, Unseen Participants, and Tail Classes; V, N, and A denote verb, noun, and action. Unseen Participants are test instances from participants absent from the training set. Tail Classes are the smallest classes whose instances account for 20\% of the training data; a tail action class has either a tail verb or a tail noun. The bold row indicates our submission.}
  \label{tab:leaderboard}
  \scriptsize
  \setlength{\tabcolsep}{2pt}
  \resizebox{\textwidth}{!}{%
  \begin{tabular}{r l r r r r r r r r r r r r}
    \toprule
    Rank & Participant & Score & PT & TL & TD
    & \multicolumn{2}{c}{Overall}
    & \multicolumn{3}{c}{Unseen Participants}
    & \multicolumn{3}{c}{Tail Classes} \\
    \cmidrule(lr){7-8}
    \cmidrule(lr){9-11}
    \cmidrule(lr){12-14}
    & & & & & & V & N & V & N & A & V & N & A \\
    \midrule
    1 & corrine & 27.95 & 2.0 & 3.0 & 3.0 & 49.22 & 52.02 & 47.83 & 57.89 & 30.90 & 42.86 & 41.53 & 21.65 \\
    \textbf{2} & \textbf{sunshinesky} & \textbf{27.91} & \textbf{2.0} & \textbf{3.0} & \textbf{3.0} & \textbf{49.19} & \textbf{51.97} & \textbf{47.83} & \textbf{57.78} & \textbf{30.93} & \textbf{42.82} & \textbf{41.46} & \textbf{21.60} \\
    3 & abardes & 25.05 & 3.0 & 3.0 & 3.0 & 49.56 & 48.57 & 46.32 & 54.43 & 26.95 & 43.48 & 38.27 & 19.16 \\
    4 & InAViT IHPC-AISG-LAHA & 23.75 & 1.0 & 3.0 & 3.0 & 49.14 & 49.97 & 44.36 & 49.28 & 23.49 & 43.17 & 39.91 & 18.11 \\
    5 & OA-OAD & 22.96 & 2.0 & 4.0 & 4.0 & 51.57 & 45.97 & 44.02 & 39.04 & 16.71 & 48.50 & 41.55 & 19.24 \\
    6 & deleted\_user\_67855 & 9.32 & 4.0 & 4.0 & 4.0 & 21.15 & 35.71 & 18.79 & 36.62 & 9.56 & 13.54 & 27.33 & 6.51 \\
    7 & AIMS\_UNICAMP & 7.51 & 1.0 & 4.0 & 4.0 & 17.81 & 21.66 & 16.47 & 20.90 & 7.13 & 9.68 & 11.21 & 4.69 \\
    8 & itruonghai1 & 5.49 & 3.0 & 4.0 & 3.0 & 26.24 & 21.87 & 20.25 & 17.25 & 3.94 & 20.01 & 15.28 & 4.11 \\
    9 & itruonghai & 5.07 & 3.0 & 4.0 & 3.0 & 35.92 & 25.37 & 28.30 & 20.06 & 4.09 & 32.04 & 20.77 & 4.58 \\
    10 & eltoncn & 4.27 & 1.0 & 4.0 & 4.0 & 14.20 & 14.30 & 13.40 & 14.91 & 3.77 & 7.98 & 7.21 & 2.61 \\
    \bottomrule
  \end{tabular}
  }
\end{table*}

\subsection{Experimental Protocol}

We evaluate TAP-JEPA with the official EK-100 action anticipation protocol.
For the final submitted model, the attentive probe replicas are optimized on the official training split plus most of the official validation split.
A small held-out subset is retained for sanity checking, qualitative inspection, and verifying the stability of the selected late-epoch window.
No hidden-test labels are used for training, model selection, or fusion-weight adjustment.

Training lasts 20 epochs.
For each exported checkpoint, the eight probe replicas are first fused into a single epoch-level prediction candidate.
The submitted file is then obtained by fusing candidates from epochs 12-20 with field-specific weights.
All outputs are written in the official action-anticipation JSON format with verb, noun, and action-pair scores.
Because most validation samples are included in the final supervised training set, we use the official hidden-test leaderboard as the main quantitative evaluation rather than reporting validation MT5R.

\begin{figure}[t]
  \centering
  \includegraphics[width=\columnwidth]{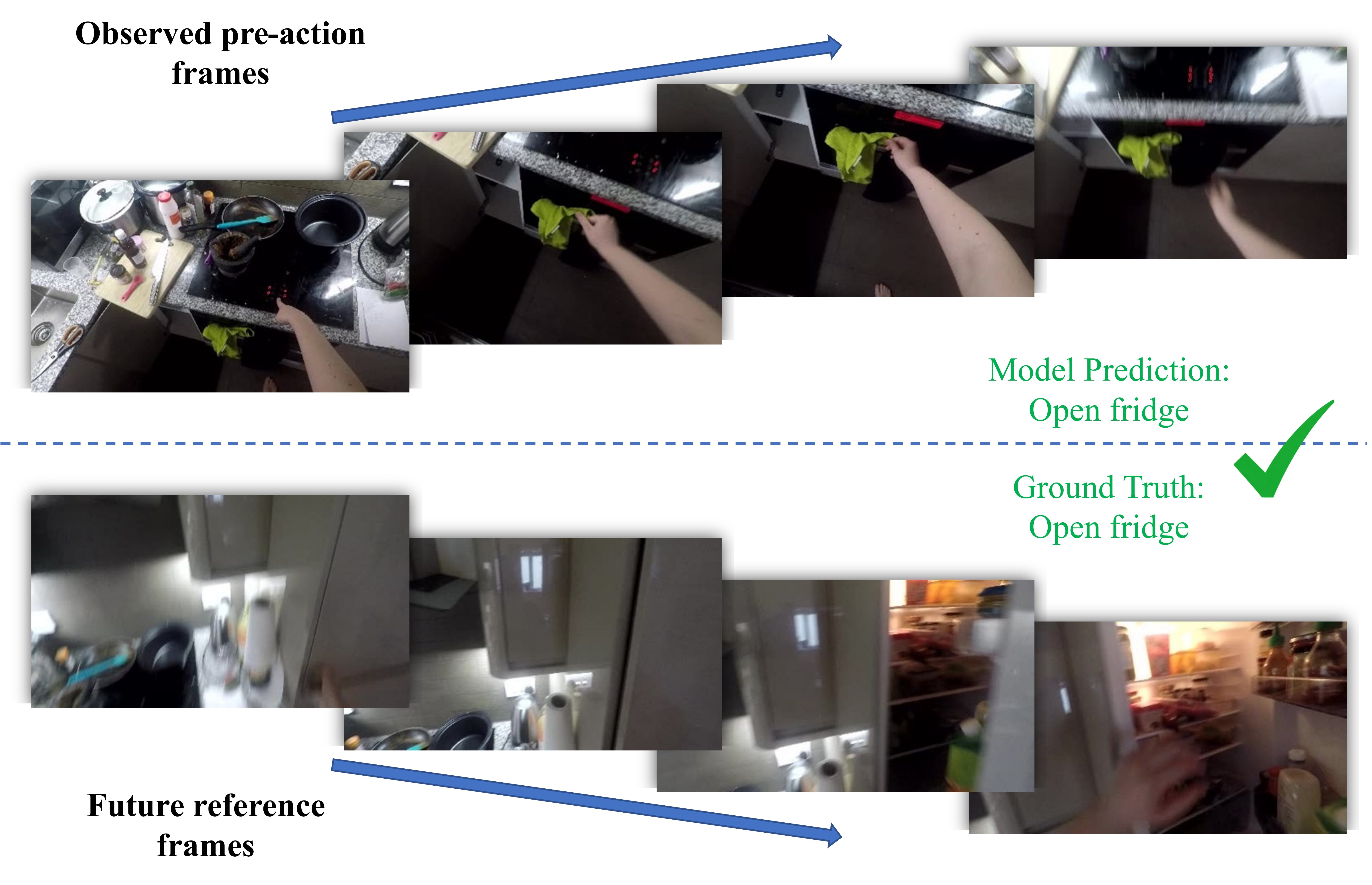}
  \caption{Validation success example. The observed pre-action frames show the camera wearer moving toward the refrigerator handle. TAP-JEPA predicts \textit{open fridge}, which matches the ground-truth action. Future reference frames are included only for explanation and are not model inputs.}
  \label{fig:case_success}
\end{figure}

\begin{figure}[t]
  \centering
  \includegraphics[width=\columnwidth]{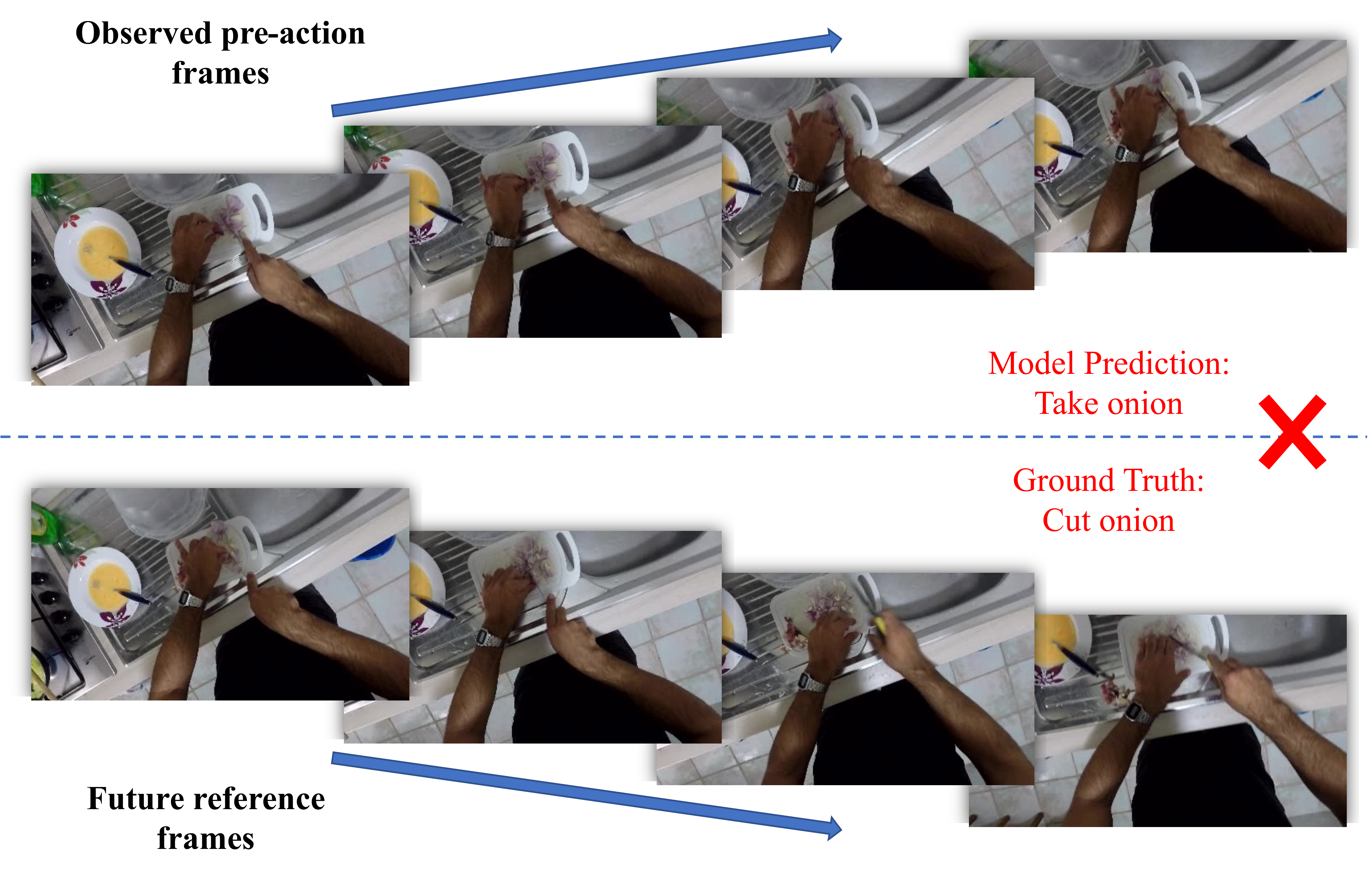}
  \caption{Validation failure example. The onion context is visible before the action, but the future manipulation is ambiguous. TAP-JEPA predicts \textit{take onion} instead of the ground-truth \textit{cut onion}. Future reference frames are included only for explanation and are not model inputs.}
  \label{fig:case_failure}
\end{figure}

\subsection{Late-Checkpoint Pool Construction}

The held-out subset is not used as a source of quantitative claims in the final report.
Instead, it is used to check that late checkpoints produce stable and complementary predictions.
In preliminary inspection, a single checkpoint can be favorable for one output field while being less reliable for another.
Therefore, the final submission does not depend on one selected epoch.

We treat epochs 1-11 as the early optimization period and use epochs 12-20 as the final candidate pool.
For every epoch in this window, replica-level scores are first aggregated across the eight independently initialized probes.
The resulting epoch-level candidates are then fused with separate weights for verb, noun, and action.
This two-stage construction keeps the ensemble compact while reducing sensitivity to individual probes and individual checkpoints.

\subsection{Hidden-Test Leaderboard Results}

Table~\ref{tab:leaderboard} reports the official open-testing leaderboard returned by the challenge server.
Our final \textit{sunshinesky} entry achieves 27.91\% overall action MT5R and ranks second.
The top entry reaches 27.95\%, so the gap is 0.04 percentage points.

The detailed metrics indicate that TAP-JEPA remains competitive beyond the overall ranking score.
It obtains 30.93\% unseen-participant action MT5R, slightly above the 30.90\% value of the first-ranked entry, and its tail-action MT5R is within 0.05 percentage points of the best value in the table.
These results suggest that frozen future-latent features and two-stage fusion are useful under participant shift and long-tailed action distributions, not only for frequent seen-participant examples.

\subsection{Qualitative Analysis}

Since hidden-test labels are unavailable, we visualize two validation examples.
The observed frames are sampled from the pre-action input window.
Future frames are shown only to clarify the ground-truth action and are never used by the model during inference.

Figure~\ref{fig:case_success} shows a favorable case in which geometry and hand motion already reveal the likely interaction with the refrigerator.
Figure~\ref{fig:case_failure} illustrates a common ambiguity in anticipation: the object is identified correctly, but the exact verb is not yet visually determined.
This contrast supports the field-wise treatment used in our ensemble, where verb, noun, and action scores are fused separately instead of being forced to share a single checkpoint mixture.

%% file: sec/4_conclusion.tex
\section{Conclusion}
\label{sec:conclusion}

We presented \textbf{TAP-JEPA}, a runner-up solution for the EPIC-KITCHENS-100 Action Anticipation Challenge at EgoVis 2026.
The method uses a frozen V-JEPA 2.1 ViT-G/384 encoder-predictor to obtain visible-context tokens and predicted future-latent tokens from pre-action clips.
A set of lightweight attentive probe replicas then predicts verbs, nouns, and action pairs with task-specific query representations.

For the final submission, the supervised probes are trained with the official training split and most of the official validation split.
Robustness is improved through two-stage score fusion: eight probe replicas are first combined within each epoch, and epoch-level candidates from epochs 12-20 are then merged with field-specific weights.
This design keeps the trainable part of the system compact while exploiting both probe diversity and late-checkpoint diversity.
Our \textit{sunshinesky} submission achieves 27.91\% overall action MT5R on the official hidden test set and ranks second in the open-testing phase, demonstrating the practical value of frozen future-predictive representations and calibrated score fusion for EK-100 action anticipation.